\documentclass[journal]{IEEEtran}

\usepackage{graphics} %
\usepackage{epsfig} %
\usepackage{mathptmx} %
\usepackage{times} %
\usepackage{amsmath} %
\usepackage{amssymb}  %
\usepackage{siunitx}
\usepackage{textcomp}
\usepackage[english]{babel}
\usepackage{csquotes}
\usepackage{subcaption}
\usepackage[font=small,labelfont=bf]{caption}
\usepackage{mathtools} %
\usepackage{algorithm}
\usepackage{algpseudocode} %
\algtext*{EndWhile}%
\algtext*{EndIf}%
\algtext*{EndFor}%
\algtext*{EndProcedure}

\newcommand{\projs}{\operatorname{proj_S}}

\newcommand{\degree}{\ensuremath{^{\circ}}}

\usepackage{tikz}
\usepackage{textcomp}
\usepackage{hyperref}
\usepackage{lipsum}

\newcommand\copyrighttext{%
  \footnotesize \textcopyright 2019 IEEE. Personal use of this material is permitted. Permission from IEEE must be obtained for all other uses. DOI: \href{<https://ieeexplore.ieee.org/abstract/document/2910739/>}{10.1109/TRO.2019.2910739}}
\newcommand\copyrightnotice{%
\begin{tikzpicture}[remember picture,overlay]
\node[anchor=south,yshift=10pt] at (current page.south) {\fbox{\parbox{\dimexpr\textwidth-\fboxsep-\fboxrule\relax}{\copyrighttext}}};
\end{tikzpicture}%
}

\newcommand{\revision}[1]{#1} %
\newcommand{\rev}[1]{#1} %

\usepackage[backend=biber,      %
    style=ieee,
    bibstyle=ieee,              %
    sortcites=true,   
    giveninits=true,            %
    backref=false
]{biblatex}
\title{\LARGE \bf
Beyond Basins of Attraction: \\ \revision{Quantifying} Robustness of Natural Dynamics
}

\author{Steve Heim, Alexander Spr{\"o}witz\\ Dynamic Locomotion Group \\ Max Planck Institute for Intelligent Systems, Germany
\thanks{Contact: heim@is.mpg.de or heim.steve@gmail.com}
}

\begin{document}

\maketitle
\copyrightnotice
\thispagestyle{plain}
\pagestyle{plain}

\begin{abstract}

Properly designing a system to exhibit favorable natural dynamics can greatly simplify designing or learning the control policy. However, it is still unclear what constitutes favorable natural dynamics and how to quantify its effect. Most studies of simple walking and running models have focused on the basins of attraction of passive limit-cycles and the notion of self-stability. We instead emphasize the importance of stepping beyond basins of attraction. We show an approach based on viability theory to quantify robust sets in state-action space. \rev{These sets are valid for the family of all robust control policies, which allows us to quantify the robustness inherent to the natural dynamics before designing the control policy or specifying a control objective.} 
\revision{We illustrate our formulation using spring-mass models, simple low dimensional models of running systems. We then show an example application by optimizing robustness of a simulated planar monoped, using a gradient-free optimization scheme. Both case studies result in a nonlinear effective stiffness providing more robustness.}

\end{abstract}

\section{Introduction}\label{sec:intro}
Animals are not only agile and efficient, but also remarkably adaptable and robust \cite{daley2006running,daley2010two}, with arguably simple control and morphology \cite{ijspeert2001connectionist, proctor2010reflexes, owaki2013simple}. \rev{Reproducing this performance in legged robots has been difficult.} Most robots use sophisticated algorithms \cite{westervelt2007feedback, koolen2016design, ponton2016convex, zhao2017robust} which rely on accurate models and state-estimation at a substantial computational cost. This reliance tends to make model-based approaches brittle. \par
Recently, there have been attempts to combine these approaches with machine learning to improve robustness and adaptability \cite{heijmink2017learning, rai2017bayesian, grandia2018contact}; however, it is notoriously difficult to apply learning directly in hardware. We are motivated by the question `how should a legged robot be designed, such that it is easier to apply model-free learning directly in hardware?'. A key part of the answer is the \emph{inherent robustness of the natural dynamics} of the system. \par
Indeed, designing a system with favorable natural dynamics can simplify the control problem \cite{sprowitz2013towards, rezazadeh2015spring, heim2016designing,ramos2018facilitating,haldane2016robotic} and enable quick learning directly in hardware \cite{tedrake2005learning,heim2018shaping}. It is, however, still unclear how to quantify and evaluate the effects of design choices on the control problem, especially in terms of robustness and ease of designing or learning the control policy. \revision{After a robot is deployed successfully, it is difficult to distinguish what is due to the mechanical design, controller design, implementation, or other factors. Designers must instead rely on experience and intuition.}\par
Many studies of \revision{natural dynamics} focus on the concept of self-stability and the basins of attraction of passively stable limit-cycles \cite{ringrose1997self,schwab2001basin,geyer2002natural,rummel2008stable} \revision{or open-loop stable limit-cycles \cite{wu20133}.}
In this study, we advocate the importance of stepping away from thinking in terms of limit-cycles and their basins of attraction. 
We present a formulation grounded in viability theory which allows us to quantify the inherent robustness of the natural dynamics, prior to specifying the control policy parametrization or control objective.

\subsection{Natural Dynamics and Spring Mass Models} \label{sub:natural}
Perhaps the clearest example of natural dynamics is Tad McGeer's passive dynamic walker \cite{mcgeer1990passive}: this purely mechanical system with no sensors or actuators (and hence no control) exhibits passively stable limit-cycles for downhill walking. This idea has been extended in several robots, adding a little actuation and control to allow walking on level ground \cite{wisse2006design,bhounsule2012design} and to increase the basin of attraction of the passively stable limit-cycle. A key concept is to \emph{exploit the natural dynamics}. The intuition behind this concept is that the control can be `lazy': if a perturbation pushes the system out of the basin of attraction, the control should guide it back in. Once the state is inside the basin of attraction, the control can allow the system to naturally evolve to the attracting limit cycle. \par
Simulation studies of idealized walking models such as the rimless wheel \cite{asano2012stability} and compass walker \cite{kuo2002energetics} have provided more understanding of McGeer's empirical results. These models also have passively stable limit cycles albeit with rather small basins of attraction. \par
For running, we turn to a different idealized model, the spring-mass model.
\revision{This simple model was initially developed by the biomechanics community to study running \cite{blickhan1989spring}, where the spring abstracts the natural compliance of the muscle-tendon system in the leg. While the effective leg stiffness depends on many factors including muscle activation, it is modeled as a constant parameter, and thus the model has no control inputs. Thus, at the level of abstraction of the model, the natural dynamics seem passive even though the system may have active control embedded in it. \par
This simple model, also called a \emph{template}, accurately predicts the overall behavior of many seemingly very different systems, called \emph{anchors} \cite{full1999templates}. Indeed, by proper parameter tuning, the spring-mass model can be used to accurately model diverse running systems, from humans \cite{maus2015constructing} to cockroaches \cite{jindrich2002dynamic}, bipedal \cite{rezazadeh2015spring} to hexapedal \cite{altendorfer2001rhex} robots.} \par

\subsection{Templates, Anchors, and Hierarchical Control} \label{sub:hierarch}
Spring-mass model templates are often used for understanding hierarchical control \cite{full1999templates} since the template and anchor division offers a natural split in hierarchy. A high-level control policy can be designed based on a template in a low-dimensional space, while a low-level control policy based on the anchor is designed in the high-dimensional space.
Thus, as long as the low-level controller enforces a template-like behavior on the system, the high-level controller design can be greatly simplified \cite{holmes2006dynamics, stephens09biped, hawkes2018design}. \par
In this hierarchical context, the term natural dynamics is always relative to the level of abstraction being considered. Indeed, to a high-level control policy, there is no distinction between which part of the system behavior is truly `passive' and which has been influenced by the low-level controller\footnote{This is equivalent to the split between agent and environment in reinforcement learning.}. \par
The template and anchor approach to hierarchical control has been used to develop various discrete-time high-level controllers: for example, the spring stiffness or landing angle of attack might be chosen once per step, but the continuous-time dynamics in between are left `passive' \cite{ghigliazza2005simply,arslan2012reactive, piovan2013two, cnops2015basin}. One result with this approach is that choosing an open-loop trajectory of landing angles of attack during flight can achieve deadbeat control without active control during stance \cite{wu20133, palmer2014periodic}. \par

\rev{While these results are impressive, they generally suffer from the curse of dimensionality: they are only tractable on the low-dimensional template models. Therefore, the high-level control relies on the overall system behaving as a simpler, lower-dimensional system. This is usually achieved through a combination of appropriate mechanical design, and a low-level controller that exposes a simpler dynamical behavior to the high-level controller. \par 
There are two common approaches to low-level controller design. On the one hand, a low-level control policy can enforce the dynamical behavior of a specific template model \cite{hutter2010slip,sentis2007synthesis,wensing2013high,poulakakis2009spring}. While this approach offers more rigorous guarantees on the behavior of the high-level system, it is also generally more difficult to implement in practice. \par
On the other hand, the low-level control policy can be designed to produce a lower-dimensional behavior without enforcing the specific template dynamics \cite{pratt2001virtual,altendorfer2004stability,renjewski2015exciting,martin2017experimental}. This approach requires further tuning of the high-level control policy, since it explicitly allows for a mismatch between the high-level model and the actual system behavior. \par
Robustness is a key indicator of how accurate a model needs to be, regardless of the approach taken: a policy that is robust will suffer less from model inaccuracies. Our main contribution is a means to quantify the robustness of the natural dynamics, prior to designing the high-level control policy, or even specifying its objective. We first illustrate the quantification on template models in a rigorous manner. We then show an example application using gradient-free optimization to find robust parameters of a low-level controller, without enforcing a specific template model. We are thus able to quantify robustness without relying on low-dimensional template models. \par }
\revision{
\subsection{Computation of Viability}\label{sub:compvi}
Our quantification relies on the concept of viability: a state is said to be viable if there exists a set of control actions that keeps it inside the viability kernel for all time \cite{aubin2009viability}. In other words, a state that starts outside the viability kernel will fail within a finite time, regardless of the control actions applied. \par
There has been much interest recently in computing viable sets and its dual, back-reachable sets \cite{maidens2013lagrangian}, for safe control verification and design \cite{liniger2017real,panagou2009viability,wieber2008viability,zaytsev2018boundaries}, and more recently safe learning of control \cite{lakatos2017eigenmodes,smitsafe}. Our contribution complements prior work by using a viability formulation to quantify robustness of the system design prior to control policy design. \par
Viability-based approaches share a common challenge: computing viability kernels relies on gridding the search space, making the general case intractable \cite{liniger2017real,bansal2017hamilton}. \par
For particular classes of systems, more efficient algorithms have been developed to find either inner or outer approximations of viable sets, which can generally be scaled to 6-10 dimensions \cite{bansal2017hamilton}. Thus, it is often beneficial to use approximations that fit these classes and dimension restrictions. Computation of viable sets is then performed on the low-dimensional approximation, which can be tracked using a hierarchical control strategy \cite{koolen2012capturability,fridovich2018planning}. \par
This matches well with the existing template and anchor paradigm commonly used in legged robotics. We will show an example application, in which we optimize the parameters of the low-level control policy to exhibit robust natural dynamics to a high-level control policy. \par}

\subsection{Notes on Terminology}
We use terminology common to the reinforcement learning community, such as actions instead of control inputs and control policies instead of controllers. We will speak of control policies \emph{sampling} an action, or the system sampling a state-action pair, to indicate the policy can be stochastic. \par
Much of the mathematics in the paper revolves around sets in different spaces. Capital letters such as $S$ denote spaces (in this case state space). Capital letters with a subscript such as $S_F$ denote a set in the corresponding space, the meaning of the subscript being explained in the text (in this case the set of failure states). \par

\subsection{Structure}
In Section \ref{sec:model} we cover the details of the two spring-mass models we examine, their dynamics, and a typical bifurcation diagram for the SLIP model. \par 
In Section \ref{sec:trans} we compute the viability kernel as well as the transition map in state-action space. We illustrate how this encompasses the bifurcation diagram, and why bifurcation diagrams are limiting once we introduce control. \par
In Section \ref{sec:robust} we introduce our definitions of robustness, and how to use this to evaluate two different designs of leg compliance prior to designing a control policy. \par
\revision{In Section \ref{sec:scaling} we show an example application, in which the quantification developed is used as the fitness function to perform gradient-free optimization of a simulated planar monopedal robot.} \par
In Section \ref{sec:conclusion} we summarize the key contributions of the paper, open questions, and our outlook.
\section{Spring-Mass Models} \label{sec:model}

\begin{figure}[tbh]
    \centering
    \includegraphics[width=1\columnwidth]{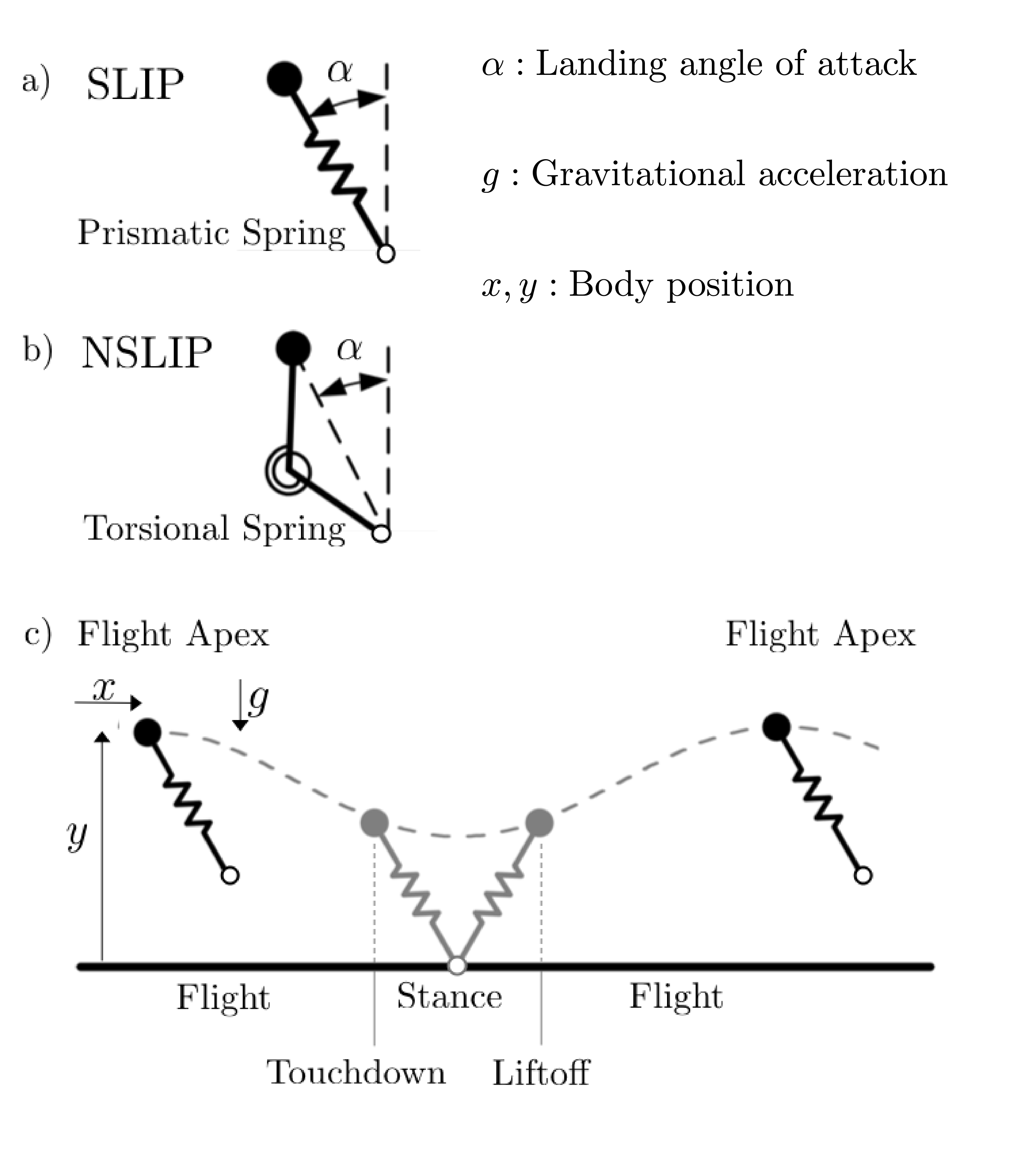}
    \caption{We focus on two spring-mass models: a) the spring-loaded inverted (SLIP) model with a linear prismatic spring, and b) a segmented leg model, with a linear torsional spring, which we will refer to as a nonlinear spring-mass (NSLIP) model. c) shows a qualitative trajectory over one cycle, starting and terminating with a flight apex event.} %
    \label{fig:slip}
\end{figure}

We use two well-studied spring-mass models to illustrate our concepts: the spring-loaded inverted pendulum (SLIP) model and a nonlinear spring mass (NSLIP) model as first studied by Rummel and Seyfarth \cite{rummel2008stable} (see Fig. \ref{fig:slip}). Both models have hybrid dynamics with the governing equations of motion switching between flight and stance phases. \par
During flight phase, the body follows a ballistic trajectory, whereas during the stance phase it follows a spring-mass motion, which depends on the modeled spring. The details of the equations of motion have been derived in \cite{blickhan1989spring,rummel2008stable}, and can be found in the appendix. For convenient comparison, we use the same parameters as in \cite{rummel2008stable}, which are similar to human averages. \rev{In this work, we consider only deterministic dynamics.}

\subsection{Discrete Analysis via Poincar\'{e} Sections}
The continuous motion of the point-mass body is fully described in \rev{planar} Cartesian coordinates by the state vector $\left[x, y, \dot{x}, \dot{y}\right]^\intercal$.
We simplify analysis by only evaluating the state on a Poincar\'{e} section at flight apex, a common approach for cyclic motion. At flight apex, potential and kinetic energy are conveniently contained in the vertical position and forward velocity respectively. Thus, the continuous state vector of $\left[x, y, \dot{x}, \dot{y}\right]^\intercal$ can be reduced to $\left[y,\  \dot{x}\right]^\intercal$. Taking advantage of the constant energy constraint, we can further reduce the system to a single state, the normalized apex height $s$, which defines our state space:

\begin{align*}
& s = \frac{E_{\text{pot}}}{E_{\text{pot}} + E_{\text{kin}}} = \frac{g y}{g y + \frac{\dot{x}^2}{2}} \\
& \text{State Space: } s \in S = \left[0, 1\right] \\
\end{align*}
where $E_{\text{pot}}$ and $E_{\text{kin}}$ are potential and kinetic energy, respectively, and $g$ is the gravitational acceleration. \par
Starting from any state at apex $s$, we can numerically integrate the continuous time dynamics until the system either transitions to a second apex height or to a failure state. We thus obtain the Poincar\'{e} map, also called a transition map, for our discrete dynamics:
\begin{equation*}
s_{k+1} = P(s_k, \alpha)
\end{equation*}
where the landing angle of attack $\alpha$ is a model parameter of interest. We use this as our control action in Section \ref{sec:trans}.\par
We will consider as failures all states in which the body hits the ground with $y=0$, as well as when the system reverses direction with $\dot{x} < 0$. More formally,
\begin{equation*}
\text{Failure Set }S_F \coloneqq \left\{ s: y = 0 \text{ or } \dot{x} < 0 \right\}\\
\end{equation*}

\subsection{Bifurcation Diagram of the SLIP Model}
A bifurcation diagram allows the study of the existence and stability of fix-points and limit-cycles, as a dependence of model parameters. \par
\begin{figure}[bth]
    \centering
    \includegraphics[width=1\columnwidth]{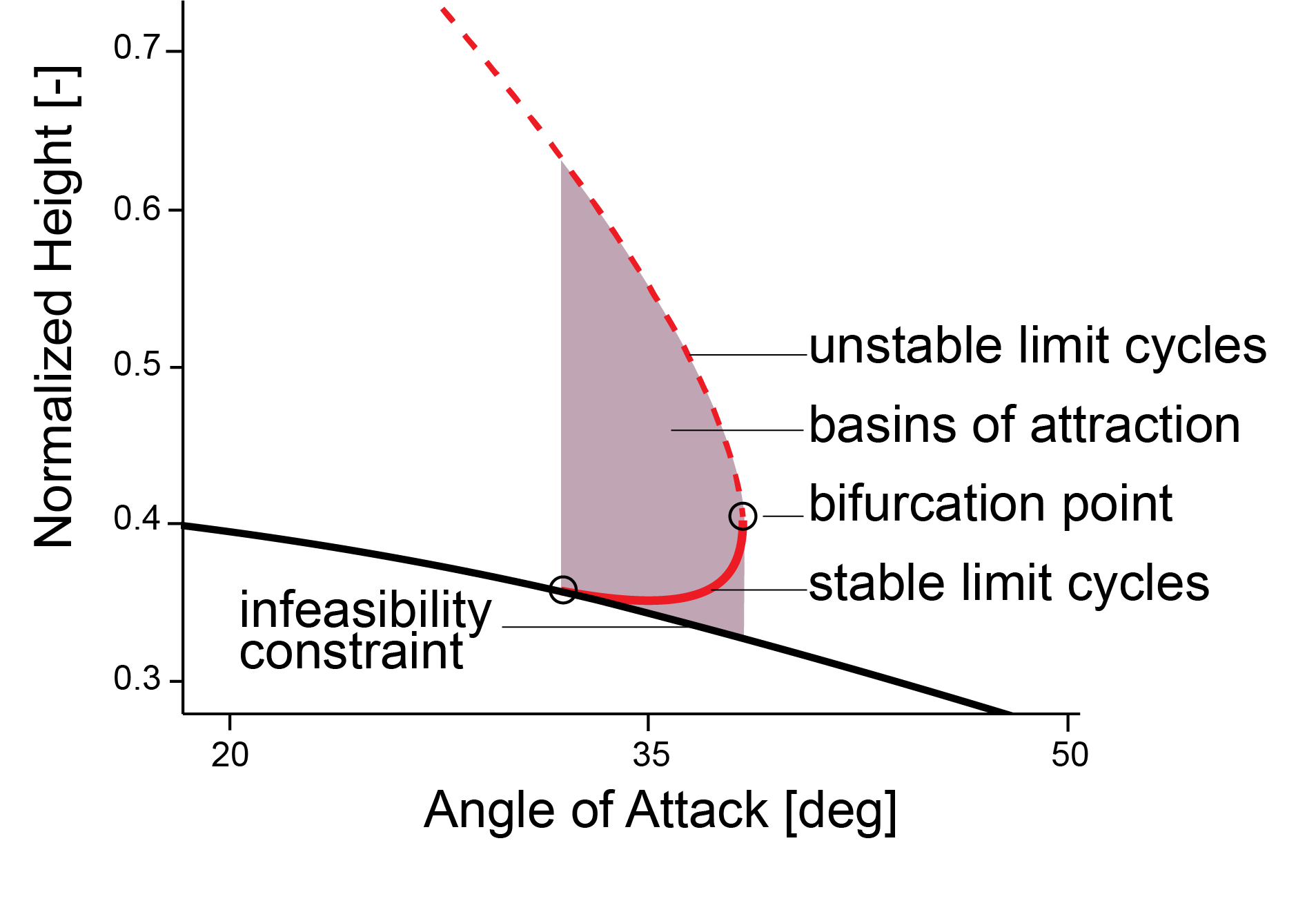}
    \caption{The bifurcation diagram of the passive SLIP model highlights the small range of parameters for which stable limit-cycles exist. The basins of attraction are bounded by infeasibility and unstable limit-cycles. Beyond these basins of attraction, however, is a lot of structure that can be exploited through control.}
    \label{fig:bifurcation}
\end{figure}

\rev{The bifurcation diagram of the SLIP model with respect to the angle of attack $\alpha$ is shown in Fig. \ref{fig:bifurcation}.} Similar bifurcation diagrams for spring-mass models can be found in \cite{merker2015stable}, and bifurcation diagrams for spring stiffness can be found in \cite{ghigliazza2005simply,rummel2008stable}. \par

We only evaluate period-1 limit-cycles, that is when $s_{k+1} = s_k$, and do not consider orbits which require multiple iterations to return to periodicity. Stable limit-cycles are marked with a solid red line and unstable limit-cycles with a dashed red line. The basins of attraction of the stable limit-cycles are highlighted by the shaded area. \par
These basins of attraction are bounded from below by an infeasibility constraint: below this line, the foot would begin underground. The unstable limit-cycles bound the basins of attraction from above: being perturbed onto an unstable limit-cycle will keep the system at that new state; beyond this threshold, it will diverge until the system fails. \par
Since either infeasibility or unstable limit-cycles bound the basins of attraction, many previous studies have been limited to identifying these bounds. The relevant range of parameters and states for studying basins of attraction tends to be narrow, as illustrated in Fig. \ref{fig:bifurcation}. We will show in the next section that there is a lot of structure outside the basins of attraction of these passively stable limit-cycles. Once we allow parameters such as the angle of attack $\alpha$ to be actively chosen as a control decision, the relevant bounds are no longer the bounds of the basins of attraction, but those of failure and viability.
\section{Natural Dynamics and Viable Control} \label{sec:trans}
We begin the section by introducing control, then evaluate the effect the natural dynamics have on the set of possible control policies. A key concept is the link between the viability kernel, a set within the state space, and the set of viable state-action pairs.
\subsection{Control Policies and State-Action Space}
We will now allow the system to choose the landing angle of attack $\alpha$ freely at each flight apex. This defines our action space $A$:
\begin{align*}
& a = \alpha \\
& \text{Action Space: } a \in A = \left[-180\degree, \, 180 \degree \right] 
\end{align*}
where $a$ is any action in $A$. In our figures we only show the relevant range, excluding the range which contains only failures or infeasible state-action pairs.\par
A control policy $\pi$ is any function that maps a state to an action, $a = \pi(s)$. As such, a policy lives in the combined state and action spaces, which we term $Q\text{-space}$\footnote{This term is chosen in reference to Q-learning in reinforcement learning.}.%
\subsection{Transition Map}
\begin{figure*}[t]
    \centering
    \includegraphics[width=1\textwidth]{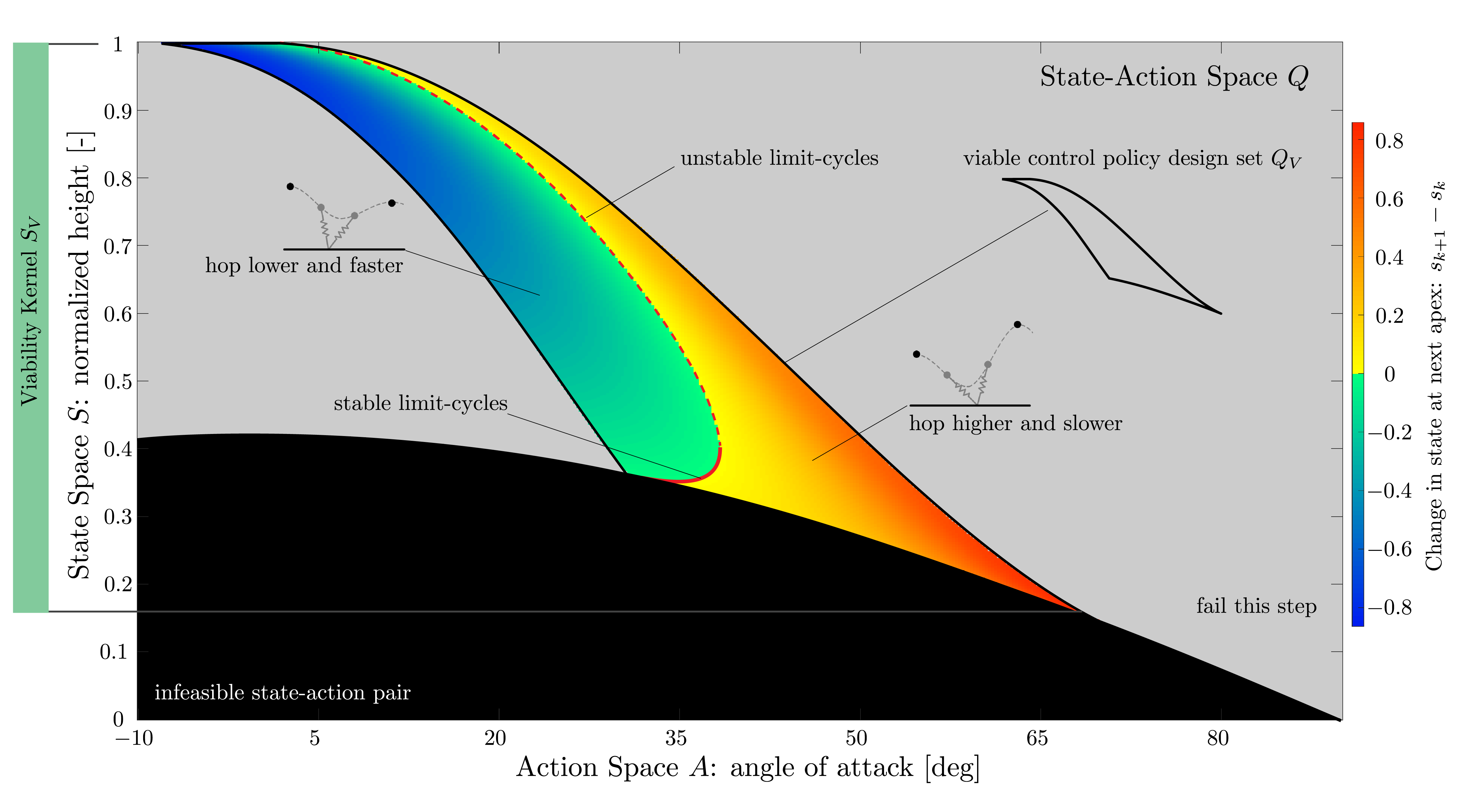}
    \caption{The lookup table of the SLIP model's transition map shows possible combinations of state (height at apex) and action (landing angle of attack), and their transition to either a second apex or a failure. State-action pairs in the gray region result in failure. State-actions in the warm and cool colored regions result in hopping higher and lower respectively, with the color indicating the change in state (vertical axis) at the next apex. Also marked are passively stable (solid red) and unstable (dashed red) limit-cycles, where the state does not change.}
    \label{fig:slipT}
\end{figure*}

\begin{figure*}[t]
    \centering
    \includegraphics[width=1\textwidth]{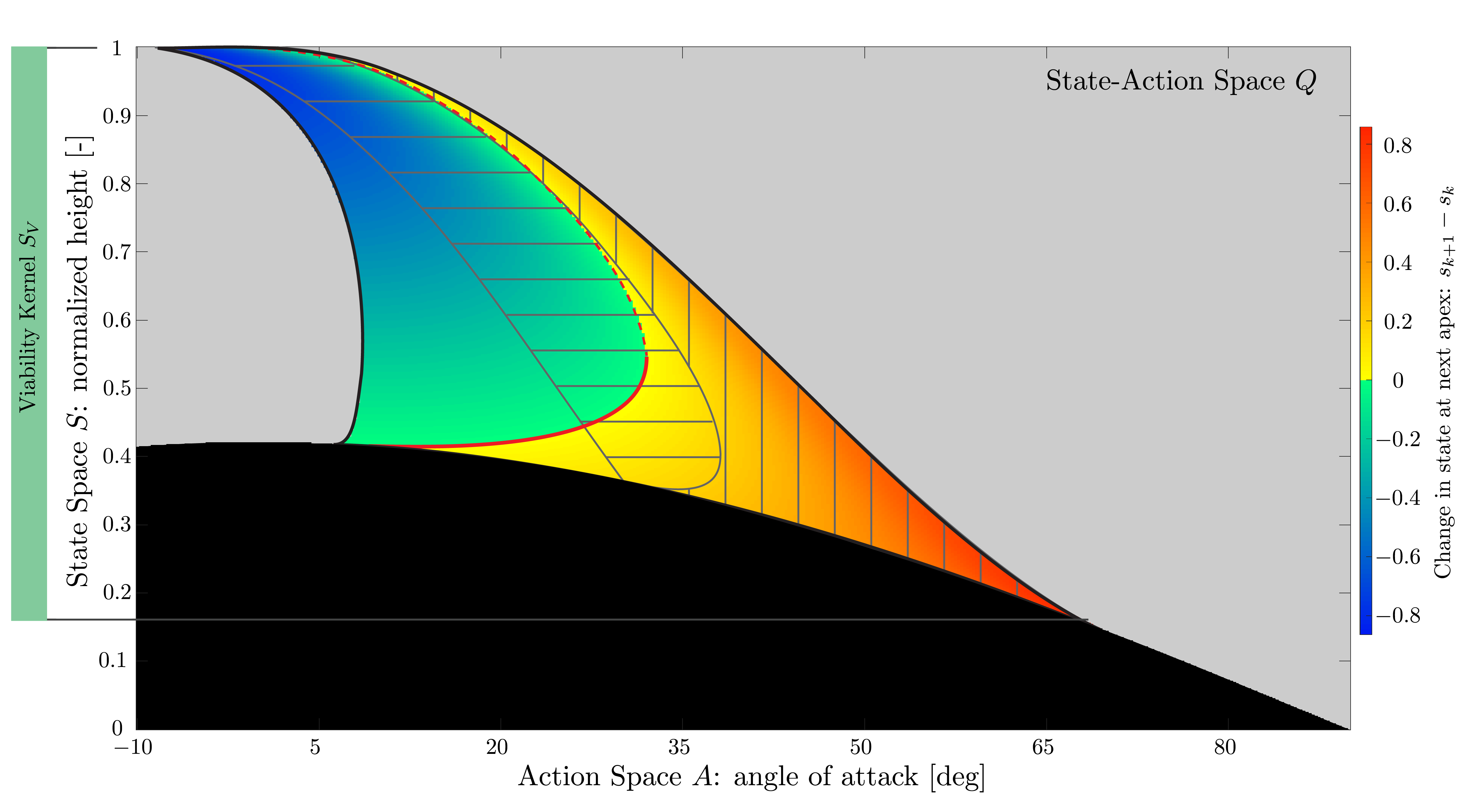}
    \caption{Although the viability kernel $S_V$ remains the same for both models, the size of $Q_V$ of the NSLIP is 36\% larger. This allows for more flexibility and robustness in designing a control policy for the NSLIP model. For reference, the $Q_V$ of the SLIP model with gray lines in horizontal and vertical for the cold and warm colored regions respectively.}
    \label{fig:rummelT}
\end{figure*}

We compute high-resolution \rev{800 by 800} grids of state-action pairs, \revision{as is commonly done for these types of problems \cite{wu20133, piovan2015reachability, cnops2015basin, zhao2017robust, zaytsev2018boundaries}.} We thus obtain a lookup table of the transition map $P(s_k, a_k)$, visualized in the state-action space $Q$ in Fig. \ref{fig:slipT} for the SLIP model and in Fig. \ref{fig:rummelT} for the NSLIP model. 

To highlight the limit-cycles, we use a color-map centered around \revision{$s_k - s_{k+1} = 0$}. The warm and cool colored regions correspond to state-action pairs that result in a higher or lower state, respectively. The gray regions are state-action pairs which result in a failure state $P(s_k, a_k) \in S_F$. \rev{The black region is composed of infeasible points in which the foot would start underground, and as such is not part of the Q-space.} \par
We call the gray region the set of failing state-action pairs $Q_F$. Its complement, the colored region, is the non-failing set of state-action pairs $Q_N$. More formally,

\begin{equation}
Q_N \coloneqq \left\{\left(s_k, a_k\right): P(s_k,a_k) \notin S_F\right\}
\end{equation}

We denote the projection of $Q_N$ onto the state space $S$ as the set $S_N = \projs(Q_N)$. Throughout the paper, we always use orthogonal projections, that is, 

\begin{equation}
\operatorname{proj_S}\left(s,a\right) = s
\end{equation}
$S_N$ is the set of controllable states, from which actions that avoid immediate failure can be selected. More formally,
\begin{align*}
S_N \coloneqq \left\{s_k : \exists \; a_k \ \text{such that} \ P\left(s_k, a_k\right) \notin S_F \right\}.
\end{align*}
\rev{The upper bound between $Q_N$ and $Q_F$ are state-action pairs that convert all kinetic energy into potential energy in one step, resulting in a state of $s = 1$. In other words, these are the equivalent of 1-step capture points \cite{koolen2012capturability}. The lower bound is a boundary to falling, meaning that the point-mass hits the ground without reaching a second flight apex.}

\subsection{Viable Sets} \label{viable transition map}

A viability kernel is the set of all states for which there is at least one time-evolution of the system which remains in the set for all time \cite{aubin2009viability}. Since all state-action pairs $(s,a) \in Q_N$ result in at least a second step, all $s \in S_N$ have at least a one failure-preventing action available. However, it is possible for a non-failing state-action pair to reach a state from which all solutions eventually reach a failed state, as was examined in \cite{heim2018learning}. In other words, there can be states from which immediate failure can be avoided, but from which the system will fail within some finite time. Thus, the viability kernel, which we will call $S_V$, is a subset of $S_N$ and \revision{the set of viable state-action pairs $Q_V$ is a subset of $Q_N$.} \par
We can compute the discretized set of viable state-action pairs $Q_V$ and its projection $S_V$ iteratively, as in Algorithm \ref{alg:viable}. \revision{In this process, we begin with an estimated $Q_V = Q_N$ and $S_V = \operatorname{proj_S}(Q_V)$. Then we alternate trimming both estimates of $Q_V$ and $S_V$: first, we check if any state action pairs $\left(s,a\right)$ in the estimated $Q_V$ maps to a state outside of $S_V$ and exclude these from $Q_V$. Then we update the estimate of $S_V$ as the projection of the new $Q_V$ estimate and repeat. If the projection does not change, each state in $S_V$ has an action available that maps back into itself and the algorithm terminates.}
\begin{algorithm}[H]
\caption{Compute Viable Sets}\label{alg:viable}
\begin{algorithmic}
\Procedure{Viable Sets}{$P, Q_N$}%
\State $Q_V \gets Q_N$
\State $S_V \gets \{\,\}$
\While{$S_V\not=\projs(Q_V)$}%
    \State $S_V \gets \projs(Q_V)$
    \ForAll {$s_{k+1} = P(s_k,a_k), \left( s_k, a_k \right) \in Q_V$}
        \If{$s_{k+1} \notin S_V$}
             \State Remove $(s_k,a_k)$ from $Q_V$
         \EndIf
    \EndFor
\EndWhile\label{computeViable}
\State \textbf{return} $Q_V,\  S_V$
\EndProcedure
\end{algorithmic}
\end{algorithm}

For the models we examine, $Q_V$ is equal or almost equal to $Q_N$ except in unusual corner cases. \par%
We can now compare the resulting $Q_V$ and $S_V$ for the SLIP and the NSLIP models (Fig. \ref{fig:robustNoiseLevels}). 
Although the set of viable states $S_V$ is the same in both models, the set of viable state-action pairs $Q_V$ is much larger for the NSLIP model. This suggests unexplored benefits of nonlinear leg compliance.

\subsection{Family of Viable Control Policies}
A control policy $\pi(s)$ must sample from $Q_N$ with non-zero probability; otherwise, it will always fail in a single step. All meaningful policies must sample from $Q_V$ with non-zero probability, or it will always fail in finite time. In order to avoid failure from every viable state for all time, a policy must sample \emph{exclusively} from $Q_V$, which we call the viable policy design space. We call the set of all such policies the \emph{family of viable control policies}. More formally, if the set $Q_V$ is non-empty, we also have a non-empty set of viable policies $\Pi_V$, where

\begin{align*} \label{eq:viable}
\begin{split}
\forall s_k \in S_V \ \exists \  \pi(s_k) \in \Pi_V,\ a_k = \pi(s_k) : \\(s_k, a_k) \in Q_V \text{ and } P(s_k, a_k) \in S_V\ \forall k
\end{split}
\end{align*}

\revision{The shape of $Q_V$ in the dimensions of $S$ and $A$ poses different constraints on the control policies $\pi(s) \in \Pi_V$ that we can design. The projection of $Q_V$ onto the dimensions of state space $S$ is the viability kernel $S_V$ itself. \par 
The volume of $Q_V$ in the dimensions of action space $A$, on the other hand, allows more flexibility in designing a viable control policy since more viable actions are available to choose from.} \par

Imagine for example a set $Q_V$ defined by a single line\footnote{A hypersurface for arbitrary dimensional state-action space.} covering all of $S$, a surjective function $f(s)$. While the viability kernel $S_V = S$ is maximal, there is exactly one deterministic control policy $\pi(s) = f(s)$ which remains viable. This can make the control policy not only difficult to design or learn, \rev{but also very sensitive to uncertainty,} as we will discuss in the next section. %
\section{Robust Natural Dynamics} \label{sec:robust}
We define robustness as the ability of a system to avoid failure in the face of uncertainty. A key objective of this work is to evaluate the robustness inherent to the natural dynamics: we care about the robustness resulting from the \revision{system} design, before specifying the policy parameterization or even the control objective (such as converging to a specific limit cycle). \par
To this end, we focus on uncertainty in action-space, in other words, the effect of noise on the control policy output.
We will use this as a basis to also examine robustness to perturbations in state-space for the family of all robust controllers. We briefly discuss the link of action noise to state-estimation noise. We do not consider model uncertainty, and leave this to future work. \par

\subsection{Computing Robust Sets}
Noise in the action space causes the system to sample a state-action pair with a different action than chosen by the policy:
\begin{align}
a & = \pi(s_k) + \eta_a \\ \label{eq:noiseAction}
s_{k+1} & = P(s_k, \, \pi(s_k) + \eta_a)
\end{align}
where $\eta_a$ is some form of noise. A robust control policy needs to ensure that the chosen output never causes the system to fail despite this noise, for all time. More formally,
\begin{align}
\begin{split}
\text{If }  & \pi(s_k) \in \Pi_R \text{ and } \eta_a \in H_a\\
\text{Then } & s_{k+1} = P(s_k, \,\pi(s_k) + \eta_a) \notin S_F \  \forall \, k
\end{split}
\end{align}
where $\Pi_R$ is the family of all robust control policies. For simplicity, we will consider noise sampled from a symmetrical bounded set $\eta_a \in H_a=\left[-\eta, \eta\right]$, where $\eta$ is some finite scalar. \par
When considering unbounded noise (such as Gaussian noise), similar arguments hold in a probabilistic sense: instead of being able to guarantee that state-action pairs allow the system to never fail, we can only guarantee that it will not fail \rev{within a finite-time horizon} with a certain probability. \par

\begin{figure*}[tbh]
    \centering
    \includegraphics[width=1\textwidth]{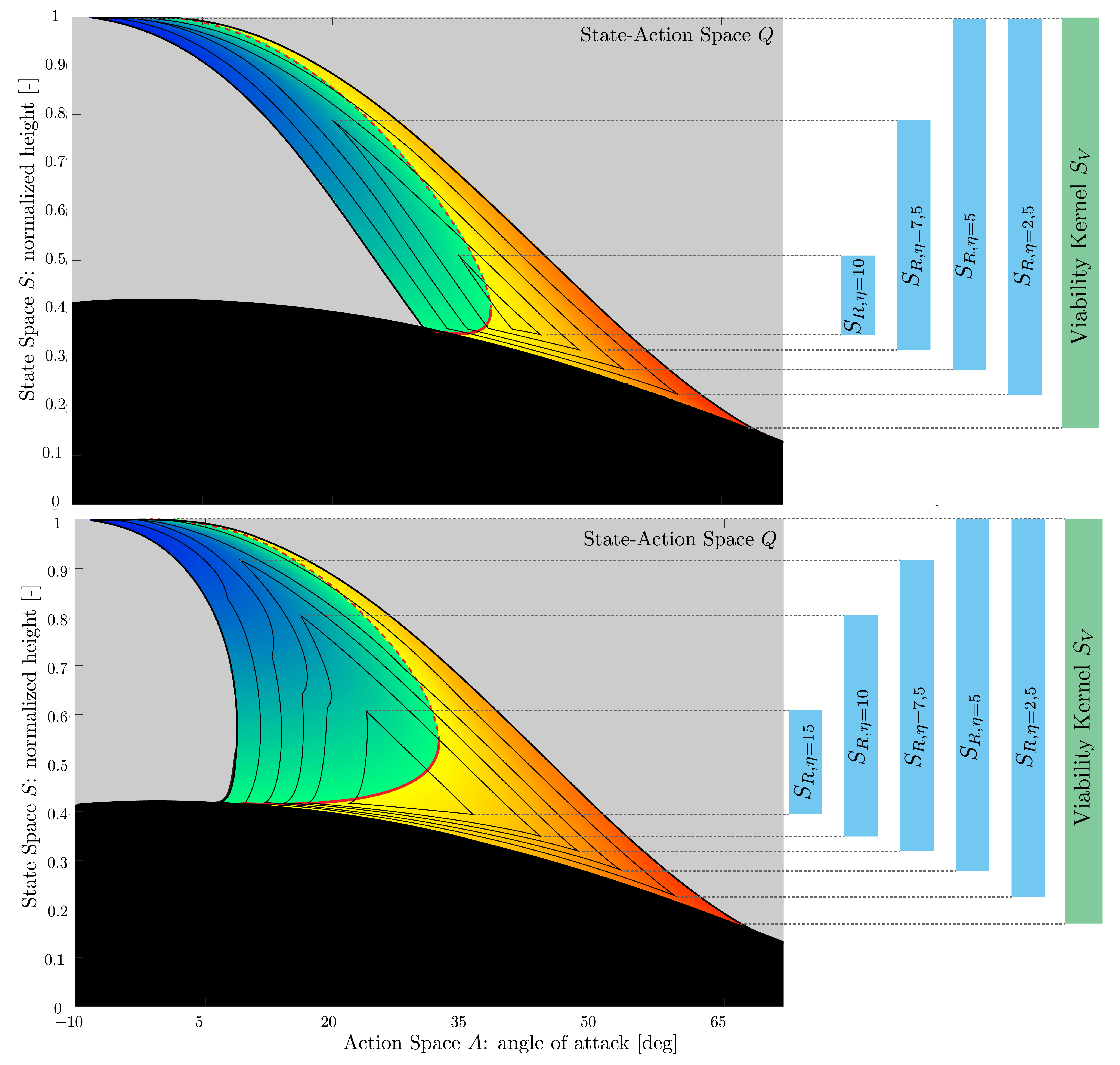}
    \caption{\revision{Robust sets for different amounts of noise are computed for the SLIP (top) and NSLIP (bottom).} The NSLIP benefits from much larger robust sets $Q_R$ for any amount of noise, which makes it easier to design or learn a robust control policy. Also, the set of robust states $S_R$ are not only larger for the NSLIP, but remain relatively large even for rather imprecise control. %
    }
    \label{fig:robustNoiseLevels}
\end{figure*}

The effect of action noise reduces the space available for controller design in two ways. First, the output of the control policy $\pi(s_k)$ must be sufficiently distant from failing state-action pairs, such that the added noise never causes an immediate failure. The second requirement is similar to that for viability: the system must always land in a state from which it can continue to sample robustly, for all time. More formally, we want that

\begin{align}\label{eq:robustness}
\begin{split}
 & s_k \in S_R \text{, }\, \pi(s_k) \in \Pi_R \text{, }\, \eta_a \in H_a:\\
 & P(s_k, \,\pi(s_k) + \eta_a) \in S_R \  \forall \, k
\end{split}
\end{align}
We call $Q_R$ the \emph{robust control policy design set}. Similar to the relation between $\Pi_V$ and $Q_V$, policies in the set $\Pi_R$ must sample exclusively from $Q_R$ in order to avoid failure for any state $s_k \in S_R$ where $S_R = \projs(Q_R)$. Such sets are shown in Fig. \ref{fig:robustNoiseLevels} for various amounts of noise $\eta$. Each of these sets is computed with the iterative process in Algorithm \ref{alg:robust}. \rev{This is essentially the same as the algorithm for computing the viable set, while also considering additional possible transitions caused by noise. Note that, if the system dynamics have certain properties, only the worst-case noise needs to be considered \cite{bansal2017hamilton}. Even without these properties, a worst-case only assumption is often sufficiently accurate in practice.}
\begin{algorithm}[tbh]
\caption{Compute Robust Sets}\label{alg:robust}
\begin{algorithmic}
\Procedure{Robust Sets}{$P,Q_V, H$}%
\State $Q_R \gets Q_V$
\State $S_R \gets \{\,\}$
\While{$S_R\not=\projs(Q_R)$}
    \State $S_R \gets \projs(Q_R)$
    \ForAll {$(s_k, a_k) \in Q_R$}      
      \ForAll {$\eta_a \in H_a$}
        \If{$(s_k, a_k + \eta_a) \notin Q_R$}
          \State Remove $(s_k, a_k)$ from $Q_R$
          \State Break
        \EndIf
        \If{$s_{k+1} = P(s_k, a_k + \eta_a) \notin S_R$}
          \State Remove $(s_k,a_k)$ from $Q_R$
          \State Break
        \EndIf
      \EndFor
    \EndFor
\EndWhile
\State \textbf{return} $Q_R,\  S_R$
\EndProcedure
\end{algorithmic}
\end{algorithm}
Importantly, the computation of $Q_R$ depends only on the set $Q_V$ and thus the set of failure state $S_F$, the transition map $P$ and the noise set $H$. It does not depend on the exact choice of policy $\pi(s_k)$, but is valid for the family of all robust control policies $\Pi_R$. In other words, we can evaluate the robustness inherent to the natural dynamics, before we design the control policy or define a control objective other than `avoid failure'.

\subsection{Evaluating Robustness of Different Legs}
We compare the robustness of the SLIP and NSLIP models for varying amounts of noise, as shown in Fig. \ref{fig:robustNoiseLevels}. \par
With the SLIP model, $Q_R$ and $S_R$ become empty sets for noise greater than $\pm 10.75\degree$, whereas in the NSLIP model the upper threshold is almost twice as large, at $\pm 20.00\degree$. 
\par
For any given amount of noise, the size of the set $Q_R$ is also much greater for the NSLIP than for the regular SLIP model.
The larger size of $Q_R$ means there is more flexibility to fulfill robustness requirements while also designing a control policy around other criteria. \par
Furthermore, action noise is one of the most common methods of introducing exploration in learning, for example with Gaussian policies \cite{kober2013reinforcement}. The amount of noise needs to be carefully balanced: more noise allows for more aggressive exploration, but it can also keep the agent from converging to the true optimum, as well as lead to unstable behaviors ending in failed states. \revision{This can be particularly troublesome for learning in hardware, requiring more samples as well as potentially damaging the robot.} Robustness to action uncertainty allows for more aggressive and effective exploration during learning. This is particularly important for applying model-free learning directly in hardware. \par
\subsection{Robustness to State Perturbations}
\begin{figure}[tbh]
    \centering
    \includegraphics[width=1\columnwidth]{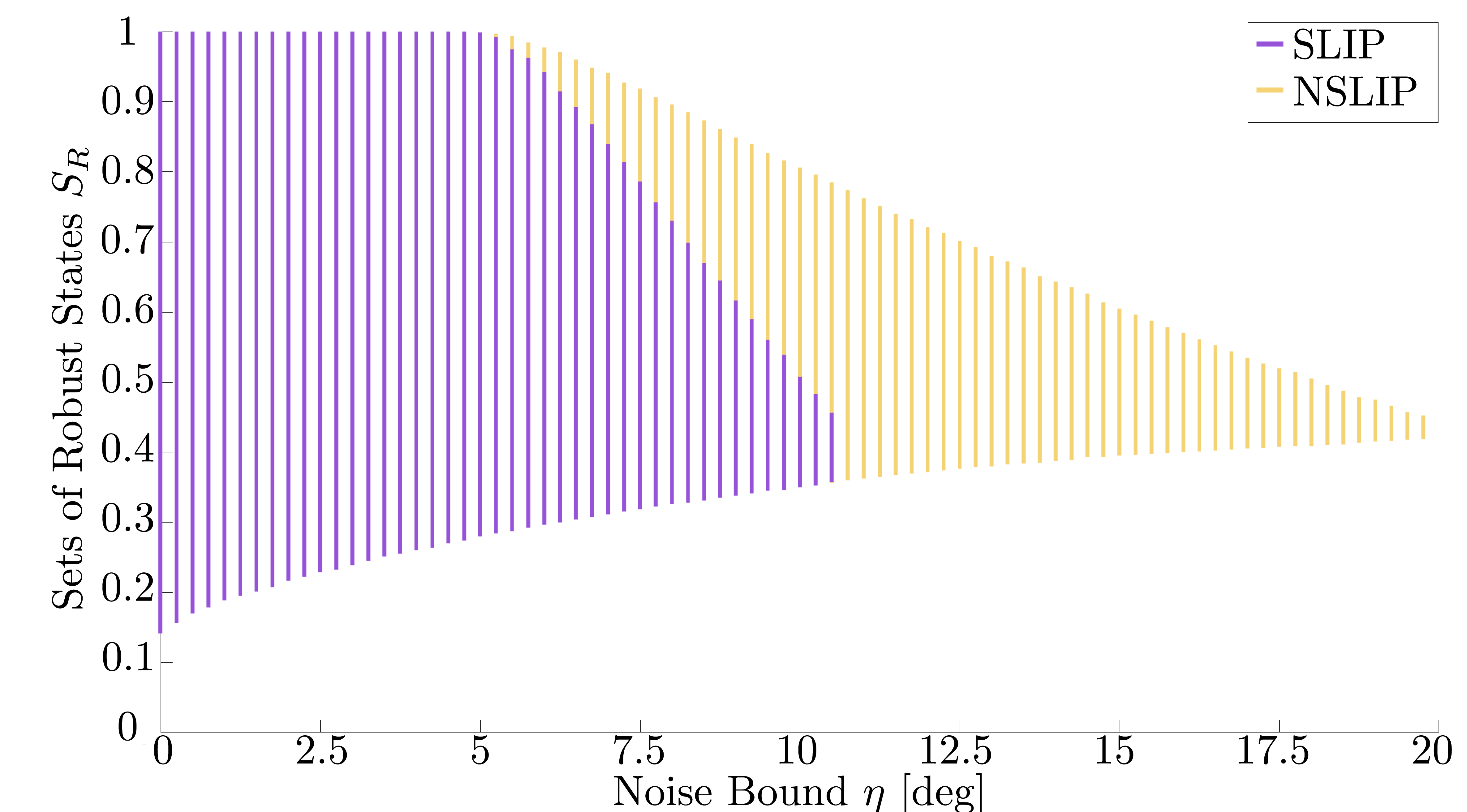}
    \caption{The size of the sets of robust sets $S_R$ remains equal for the SLIP and NSLIP models for noise bounded to less than $5 \degree$. For greater amounts of noise, the sets shrink much more rapidly for the SLIP model.}
    \label{fig:setsize}
\end{figure}

The projection of the robust policy design set onto state-space, $S_R = \projs(Q_R)$, is the set of robust states, from which any robust policy $\pi \in \Pi_R$ can always recover. Interestingly, with small amounts of noise up to $\eta < 5 \degree$, $S_R$ remains the same for both the SLIP and NSLIP models (see Fig. \ref{fig:setsize}). For greater amounts of noise, it shrinks much more rapidly for the SLIP model.\par
The set $S_R$ is particularly useful for choosing the specific control objective. For example, if we expect perturbations in state-space to have a symmetrical distribution, we would want to stabilize a limit-cycle near the center of $S_R$. On the other hand, if we expect a specific type of perturbation to occur more frequently, we can choose a limit cycle with a larger margin in that specific direction. \par
\begin{figure}[htb]
    \centering
    \includegraphics[width=1\columnwidth]{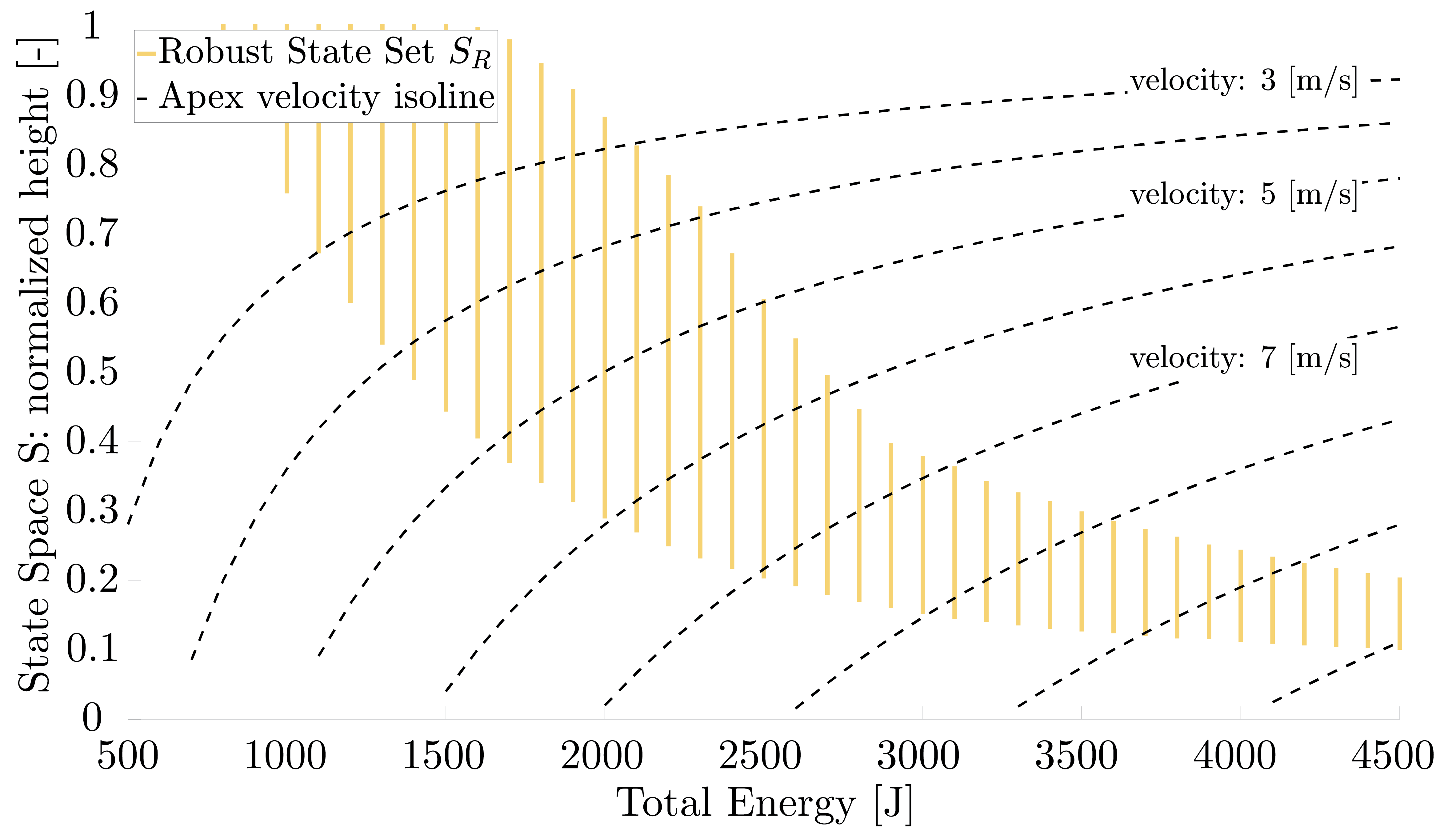}
    \caption{We show here the $S_R$ for different amounts of total energy for the NSLIP model, with noise fixed at $\eta = 7.5\degree$. For a change in ground height, the system state travels along the forward velocity isolines (dashed black). For reference, the author runs recreationally at roughly $3.2 \left[m/s\right]$, Eliud Kipchoge ran the Breaking2 marathon event at roughly $5.8 \left[m/s\right]$ and Usain Bolt holds the 100 meter dash world record at roughly $10.8 \left[m/s\right]$. The simulations shown in other graphs are all for the fixed energy level of 1'860 Joules.}
    \label{fig:stepdown}
\end{figure}

As a specific example, a well-studied state perturbation is a change of ground height between steps \cite{daley2006running,sprowitz2013towards,wu20133}. This type of perturbation involves a change in total energy: the forward velocity at apex remains the same, though the effective height (and thus potential energy) changes. We can compute $S_R$ at different energy levels to then pick out operating points that remain robustly controllable across different energy levels, as shown in Fig. \ref{fig:stepdown}. Assuming symmetric distribution of perturbations, the control objectives should be chosen to maximize the distance from the edge of the viability kernel in each direction. For a given desired forward velocity, we can thus choose a total energy that centers the normalized height to perturbation along the vertical axis (constant energy perturbation) and along the forward velocity isolines (ground height change).

\subsection{Robustness to State Estimation Uncertainty}
Sensory noise causes the control policy to sample an action based on a noisy estimate of the state:
\begin{align}
a = \pi(s + \eta_s) \label{eq:noiseState}
\end{align}
where $\eta_s$ is the noise in state space.
There is an equivalence between $\eta_s$ and $\eta_a$: the action used deviates from what a control policy would determine under perfect conditions, whether this is due to noise in action space or state estimation. This equivalence can be directly calculated using eq. \ref{eq:noiseAction} and eq. \ref{eq:noiseState}:

\begin{align*}
\begin{split}
\pi(s) + \eta_a = \pi(s + \eta_s) \\
\eta_a = \pi(s + \eta_s) - \pi(s)
\end{split}
\end{align*}
If the control policy $\pi$ is affine, the equivalence is trivially $\eta_a = \pi(\eta_s)$ and for bounded estimation noise $\eta_s$ the equivalent action noise $\eta_a$ is also bounded. Otherwise, we cannot guarantee bounds are available.
Since this equivalence is dependent on the specific control policy, we do not investigate it further here. Suffice it to say, increasing robustness to action uncertainty can only improve robustness to state-estimation uncertainty as well.

\revision{\subsection{Model Comparison}

Previous studies of spring-mass models by Rummel and Seyfarth and others \cite{rummel2008stable,owaki2006enhancing,karssen2011running} have suggested that nonlinear effective leg compliance can improve stability. These studies focus on finding basins of attraction with a fixed parameter set. As such, they focus specifically on limit-cycle motion and only provide insight to robustness to state perturbations. \par
With their numerical studies, Rummel et al. show that, compared to a linear leg compliance, a nonlinear leg compliance has a broader range of parameters which exhibit passively stable limit-cycles. These limit-cycles also tend to have larger basins of attraction. However, at higher velocities, the model with nonlinear spring stiffness no longer exhibits passively stable limit cycles, whereas with a linear spring this property is retained. These results suggested that nonlinear compliance is only beneficial at lower running speeds \cite{rummel2008stable}. \par
Using our formulation, we can evaluate robustness to state perturbations not only for an open-loop system but for any robust control policy. Our results confirm that, even with a maximally robust control policy, the set of robust states $S_R$ shrinks at higher speeds (see Fig. \ref{fig:stepdown}), though not as drastically as the basins of attraction studied by Rummel et al. \par}
\revision{
\section{Optimizing Natural Dynamics for Robustness} \label{sec:scaling}
\revision{
As an example application, we use our quantification to optimize the robustness of a simulated planar monoped with a 2-segment leg with a hierarchical control structure, shown in Fig. \ref{fig:monoped}. The kinematic tree of the simulated system matches a robot testbed we currently use in our lab, though we have adjusted the parameters to be consistent with \rev{the models in the previous section}. The system consists of three links: a floating-base free to move in the plane, but without rotation, and a two-link leg. Both hip and knee joints are actuated, resulting in an 8-dimensional state space and a 2-dimensional action space. Rigid impacts and ground-reaction forces are solved as described in \cite{remy2011matlab, hutter2010slip}.} \par
\begin{figure}[tbh]
    \centering
    \includegraphics[width=1\columnwidth]{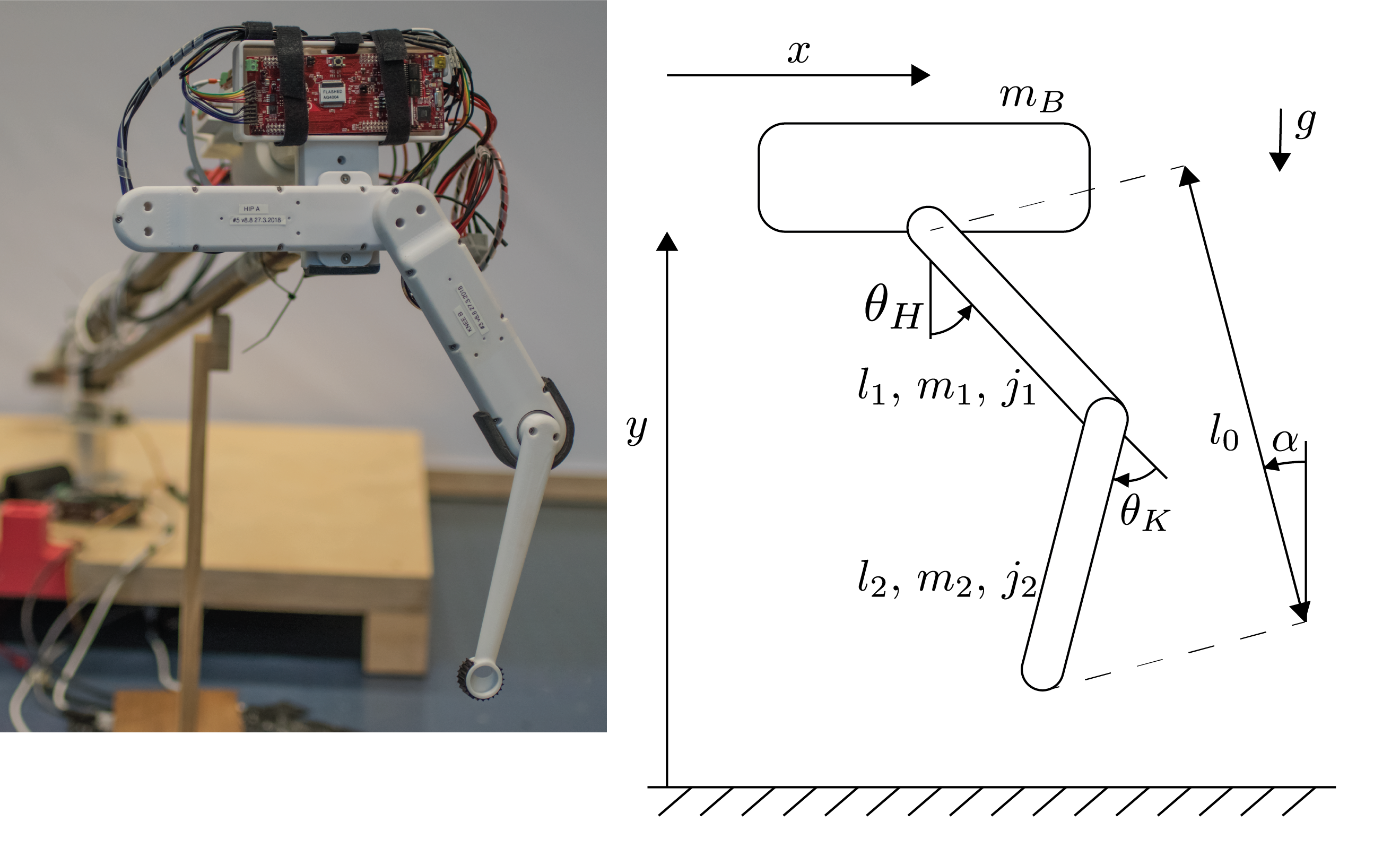}
    \caption{\revision{The simulated system is based on a hardware testbed, which is rigidly attached to a boom. Thus the floating base is limited to two degrees of freedom. Two additional degrees of freedom, the hip and knee joints, are both actuated. Thus the system has 4 position coordinates $q = \left[x, y, \theta_H, \theta_K \right]^\intercal$, an 8 dimensional state space $\left[q, \dot{q}\right]^\intercal$ and a 2 dimensional action space $\left[\tau_H, \tau_K\right]^\intercal$, where $\tau_H$ and $\tau_K$ are the hip and knee torques, respectively. The robot shown is designed by our colleague Felix Grimminger.}}
    \label{fig:monoped}
\end{figure}
\rev{We use the volume of the robust set $Q_V$ as the fitness function for a particle swarm optimization (PSO), a standard gradient-free optimization scheme. Thus, instead of requiring the low-level controller to enforce a specific template model, we improve its robustness in a general sense. The resulting natural dynamics allow for a high-level control policy to be implemented more reliably.}

\subsection{High-Level State-Action Space}
\revision{The choice of the high-level state-action space is based on the spring-mass models and classic Raibert control \cite{raibert1986legged}, which share many similarities. The structure is shown in Fig. \ref{fig:hierarch}.}

\begin{figure}[bth]
    \centering
    \includegraphics[width=0.9\columnwidth]{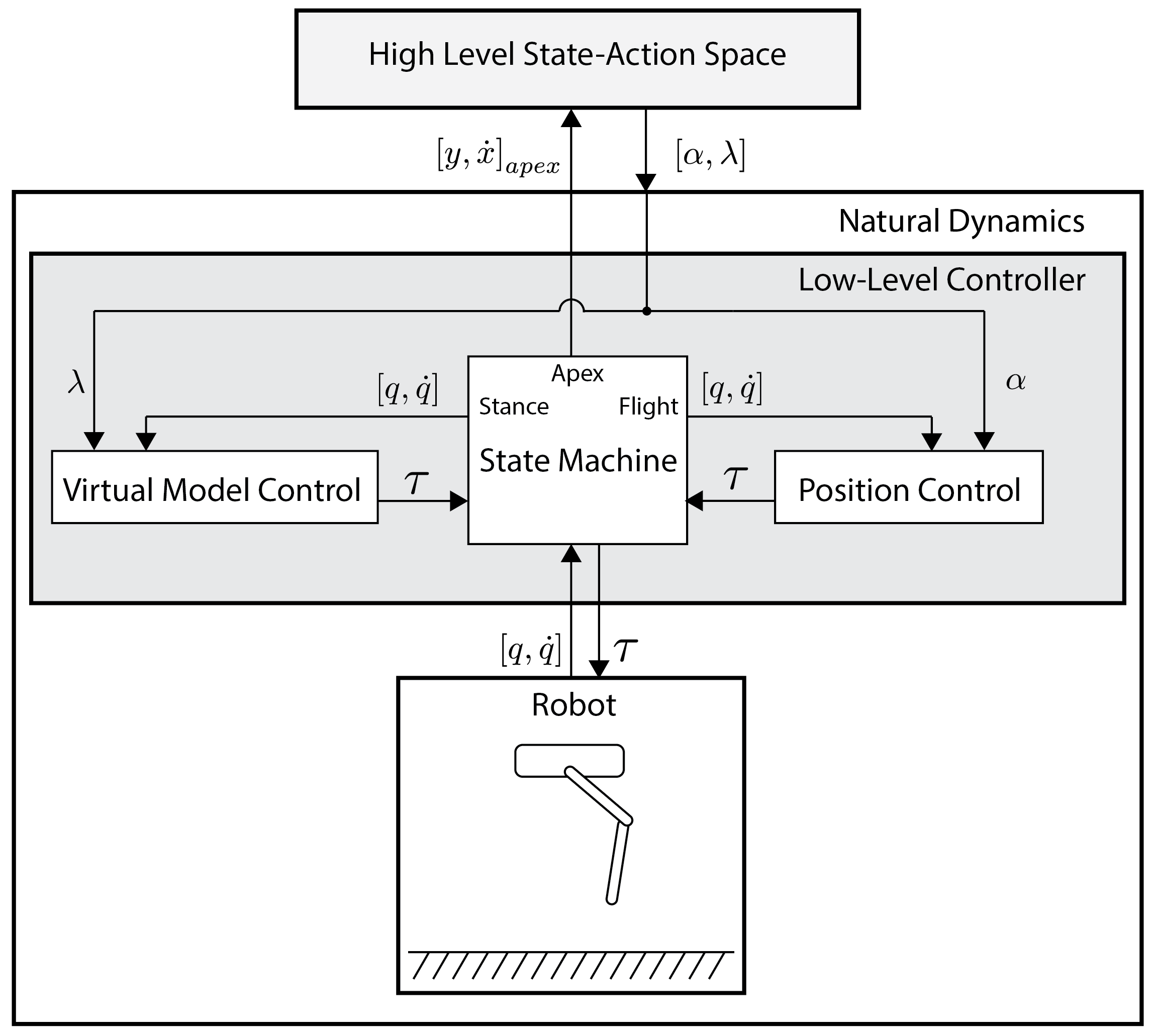}
    \caption{\revision{The high-level state-action space is composed of the height and forward velocity \rev{of the floating base} at apex $\left[y, \dot{x}\right]^\intercal_{apex}$, the desired landing angle of attack $\alpha$ and the thrust factor $\lambda$. The natural dynamics considered are those relative to the high-level. These include both the rigid-body dynamics of the simulated robot as well as the embedded low-level controller.}}
    \label{fig:hierarch}
\end{figure}
\revision{The state is defined on the Poincar\'{e} section at flight apex, as introduced in Section \ref{sec:model}. Since the system is not energy-conservative, both the height and forward velocity \rev{of the floating base} at apex must be considered, resulting in the state vector $\left[y, \dot{x}\right]^\intercal_{apex}$. \par
The action space is defined as a desired landing angle of attack $\alpha$, constrained within $0$ and $45\degree$, and a thrust factor $\lambda$ applied during stance, constrained within $1$ and $2$. This results in a 4-dimensional state-action space in the high-level, which is amenable to direct computation of a sufficiently dense grid. \par
Although our choice of the state-action space is largely motivated by Raibert control, we make no restrictions on the high-level control policy and do not decouple the states and actions. }

\subsection{Low-Level Controller}
\revision{
The low-level controller is a state-machine that switches between flight and stance. \par
During flight, a standard PD position controller tracks the desired landing angle of attack $\alpha$ dictated by the high-level control policy. 
The resting length of the virtual leg, $l_0$, is set as a constant parameter less than the maximum leg length to avoid reaching singularities. Thus $\alpha$ uniquely determines the desired foot position during flight. Since there are two possible joint configurations for each desired foot position, this orientation is also set as a constant parameter in the computation of the inverse kinematics. Thus $\alpha$ also uniquely determines the desired joint angles. \rev{During the first flight phase, from apex till touchdown, $\alpha$ is freely chosen as the action. For the second flight phase, from liftoff till the next apex, $\alpha$ is reset to the default position $0$. Thus the initial leg configuration at each apex is expected to be the same. \par
During stance\rev{, we do not enforce the dynamics of a spring mass template model. Instead, compliant} behavior is achieved via virtual model control (VMC) \cite{pratt2001virtual,renjewski2015exciting}. Torques are computed to mimic a relatively arbitrary leg compliance:
\begin{equation} \label{eq:lowlevel}
\left[\tau_H, \tau_K \right]^\intercal = \begin{cases}
B J_c^\intercal\, k_v\Delta l\, + \, K_j\Delta\theta & \text{if } \dot{y} < 0\\
\lambda\left(B J_c^\intercal \ k_v\, \Delta l\, + \, K_j\Delta\theta \right) & \text{otherwise}\\
\end{cases}
\end{equation}

where $\left[\tau_H, \tau_K \right]^\intercal$ are the hip and knee torques, $B$ is the actuator selection matrix, $J_c$ is the contact Jacobian, $k_v$ is the stiffness coefficient of a virtual linear spring between hip and foot, $\Delta l$ is the deflection of the virtual leg from rest, $K$ is a symmetric linear matrix, and $\Delta\theta$ is the joint deflection of the leg from rest. The diagonal coefficients of $K$ can be interpreted as virtual springs on the corresponding joints, while the off-diagonal coefficient serves as a mixing term. As long as $K$ is positive-definite, $K$ results in a nonlinear compliance with respect to the virtual leg deflection $\Delta l$.
In similar fashion to classic Raibert control \cite{raibert1986legged}, additional thrust is triggered once the body reverses direction by amplifying joint torques by the thrust factor $\lambda$, as dictated by the high-level control policy. \par
We assume accurate tracking of $\alpha$ during flight phase, which is achieved through proper tuning of the PD gains. This is important to ensure well-behaved high-level dynamics for two reasons. First, to ensure that each high-level state-action pair results in a unique state at touchdown. Second, to ensure that the robot leg returns to the same resting configuration at each apex. In this manner, the leg masses can be lumped with the floating base to determine potential and kinetic energy, meaning that the high-level state $\left[y, \dot{x}\right]^\intercal_{apex}$ fully describes the system energy. The transition map $P$ thus provides a unique map for each high-level state-action pair, and the viable sets $S_V$ and $Q_V$ can be directly computed in the high-level state-action space.}}

\subsection{Optimization Setup}
\revision{We use a standard PSO implementation based on \cite{clerc2002particle}. The parameters optimized are the stiffness coefficients of the virtual leg in the low-level stance controller, $\left[k_v, k_{11}, k_{22}, k_{ij}\right]$, where $k_{11}$ and $k_{22}$ form the diagonal of the symmetric matrix $K$, and $k_{ij}$ is the off-diagonal term. \par
As fitness function, we choose to maximize the \rev{hypervolume} enclosed by the viable set $Q_V$ in the high-level state-action space. For our systems, we have found that maximizing the hypervolume of $Q_V$ and $Q_R$ generally leads to the same results for reasonable amounts of noise. \rev{Each dimension of the state is normalized by heuristically determined bounds on maximum height and forward velocity, and the dimensions of the action space are bounded by their corresponding constraints. The hypervolume is calculated by summing and then normalizing the points inside the set. Thus, a fitness of 1 means that for any state, all actions are viable. A fitness of 0 means that for any state, all actions are outside the viable set.} \par
For the results shown, 25 particles were initialized at random. Convergence tolerance on the fitness variance was set to $10^{-5}$, which was reached after 12 iterations, taking roughly 3.5 hours on a 28-core desktop. During the optimization, we used a low-resolution grid with 160'000 points to speed up computation. Note that a lower resolution will result in a more conservative estimate of the sets, but not in mislabeled points in the set. The simulation parameters used are:}

\begin{center}
\begin{tabular}{|l l l l|} 
\hline
\multicolumn{4}{|l|}{\textbf{Mechanical Parameters}} \\
\hline
gravitational acceleration &$g$ :& 9.81 &$\left[m/s^2\right]$ \\
\hline
body mass &$m_B$ :& 65 &$\left[kg\right]$ \\
\hline
upper leg length &$l_1$:& 0.5 &$\left[m\right]$ \\
\hline
upper leg mass &$m_1$:& 10 &$\left[kg\right]$ \\
\hline
upper leg inertia &$j_1$:& 2 &$\left[Kgm^2\right]$ \\
\hline
lower leg length &$l_2$:& 0.5 &$\left[m\right]$ \\
\hline
lower leg mass &$m_2$:& 5 &$\left[kg\right]$ \\
\hline
lower leg inertia &$j_2$:& 2 &$\left[Kgm^2\right]$ \\
\hline
\multicolumn{4}{|l|}{\textbf{Low Level Control Parameters}}  \\
\hline
leg resting length &$\l_0$ :& 0.85 &$\left[m\right]$\\
\hline
saturation torque &$\tau_{max}$:& 2000 &$\left[Nm\right]$\\
\hline
Hip joint PD gains &$\left[k_p, k_d\right]$:& $\left[500,50\right]$ &$\left[-\right]$\\
\hline
Knee joint PD gains &$\left[k_p, k_d\right]$:& $\left[500,25\right]$ &$\left[-\right]$\\
\hline
\end{tabular}
\end{center}

\subsection{Optimization Results}
\revision{
We compare the robustness with a virtual leg compliance roughly matching that of the SLIP model, with stiffness coefficient $\left[k_v, k_{11}, k_{22}, k_{ij} \right] = \left[8, 0, 0, 0\right]10^3$, versus one with the stiffness coefficients resulting from the optimization, $\left[k_v, k_{11}, k_{22}, k_{ij} \right] = \left[8.1, 5.0, 0.9, -0.5\right]10^3$. The viability kernels $S_V$ are visualized in Fig. \ref{fig:monokernel}. The intensity of the color-map indicates the portion of the action space which is viable for each point in state space. The red triangle marks an arbitrary operating point, $\left[y,\  \dot{x}\right]^\intercal = \left[1, 1\right]^\intercal$, and the action space for this state is shown in the image inset. In the action space, the action-pair leading to limit-cycle motion is also marked by a red triangle. To illustrate improved robustness, 50 actions are uniformly sampled around the operating point assuming bounded noise $\eta = \left[5\degree, 0.1\right]^\intercal$ (orange circles) and an additional 50 with bounded noise between $\eta$ and $2 \eta$ (blue circles). \par
As in the comparison between the SLIP and NSLIP models in the previous section, the viability kernel $S_V$ in state space remains nearly identical for both systems. The volume of the set of viable state-action pairs, however, increases from 0.08 to 0.23, over 2.8 times. The noisy sampling of actions around the operating point shows the decreased sensitivity to action noise with the optimized nonlinear compliance.
\begin{figure*}[tbh]
    \centering
    \includegraphics[width=1\textwidth]{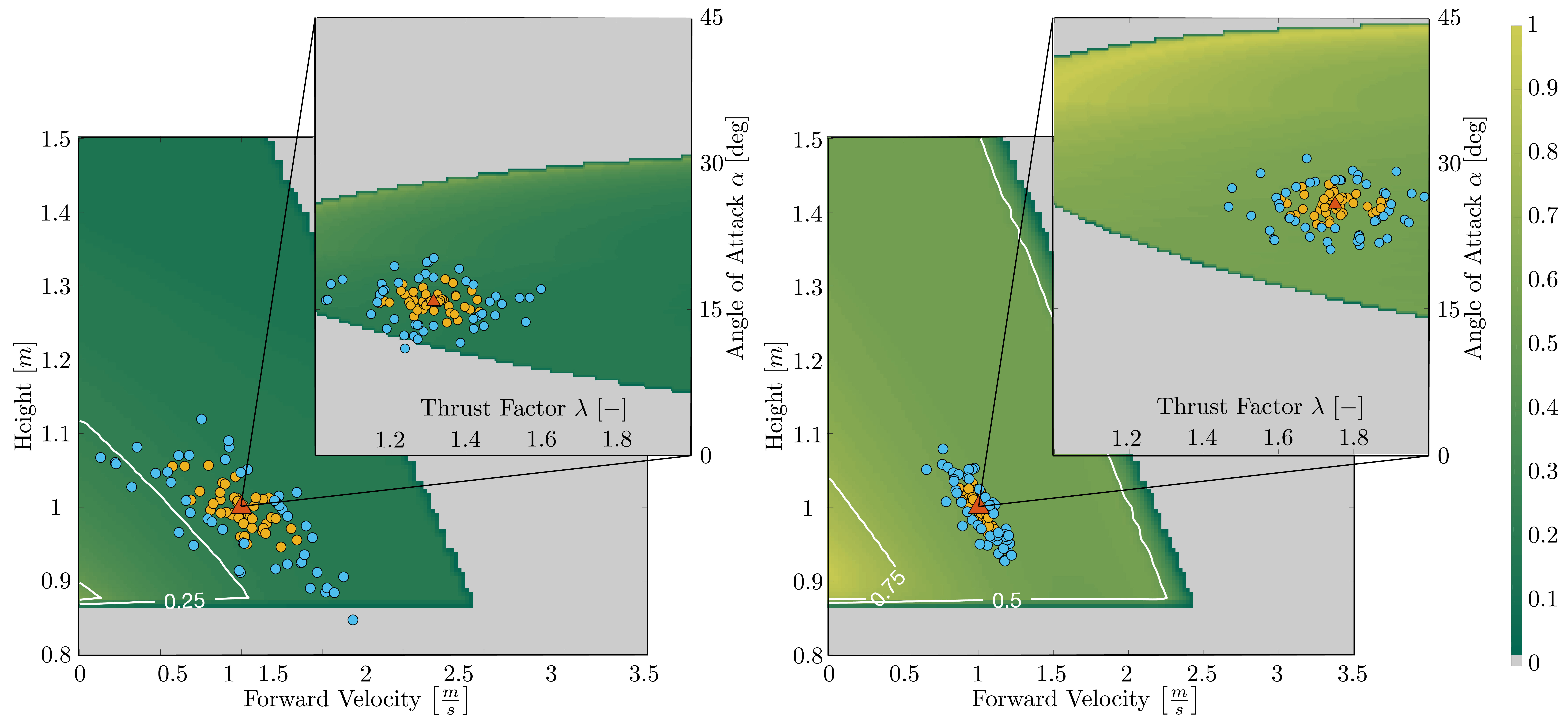}
    \caption{\revision{Shown are the viability kernels in the high-level space for the initial monoped (left) and after optimizing virtual compliance (right), along with the action-space for the operating point $\left[y,\  \dot{x}\right]^\intercal = \left[1, 1\right]^\intercal$, shown in the image insets. A red triangle marks the state-action pair which leads to limit-cycle motion on the operating point, in both the state and action spaces. In the action space, the orange and blue circles mark actions randomly sampled around the operating point and with bounded noise $\eta = \left[5\degree, 0.1\right]^\intercal$, and between $\eta$ and $2\eta$, respectively. The states reached by these state-action pairs are marked with their respective colors in the state-space, which shows the much lower sensitivity experienced by the optimized monoped. The intensity of the color-map indicates for each point in state space, the portion of the action space which is viable. In the action space (image insets), the color-map indicates the intensity of the state that would be reached if that state-action pair were sampled.}}
    \label{fig:monokernel}
\end{figure*}
\rev{In Fig. \ref{fig:monokernel} we chose an arbitrary operating point for the sake of simplicity and fair comparison. In practice, an operating point can be chosen based on the robustness of that point in state-space. Conversely, instead of optimizing the overall robustness of the system, the fitness function can be weighted to bias robustness near a predetermined operating point.}
}}

\section{Conclusion and Outlook} \label{sec:conclusion}
\revision{We have presented a formulation for computing viable and robust sets in state-action space which allows the inherent robustness of a system to be quantified, prior to specifying the control policy parameterization or objective. Different system designs can thus be compared quantitatively. \par
We have illustrated this formulation on the spring-mass model, a low-dimensional system commonly used to synthesize control strategies for running robots. Furthermore, we have shown an example application using our quantification to perform gradient-free optimization. The system optimized is a simulated planar monoped with a two segment leg and a hierarchical control structure. The low-level controller parameters are optimized to improve robustness of the natural dynamics, as relative to the high-level state-action space.
 \par
\rev{An important advantage of this formulation is that the natural dynamics robustness can be optimized without enforcing the dynamics of a specific template model, which is often challenging and requires extensive tuning, developing accurate models as well as state estimation \cite{sentis2007synthesis,herzog2016momentum,flayols2017experimental}. Instead, the inherent robustness will allow control policies designed on simple model abstractions to be leveraged despite inaccuracies.} \par
To the best of our knowledge, prior work in viability theory focuses on evaluating robustness of a specific control policy, or on synthesizing control policies directly, and computation is limited to viability kernels in state space. \par
The notable exception is the work of Zaytsev et al. \cite{zaytsev2018boundaries}, which also computes viable sets in state-action space. Aside from the minor difference in studying walking instead of running models, Zaytsev et al. focus on the connection between controllability and viability. This is used to qualify how robust a given control policy is, how appropriate different templates may be for a given control task and given robot, and to motivate the statement that planning two steps ahead is sufficient. While we use the same state-action space formulation, we take a different approach to quantification by evaluating bounded noise in action space, which is more suitable for our motivating question: how to design natural dynamics that are easy to exploit? Indeed, we show why this is the only type of uncertainty which can be considered for the family of all robust control policies, without setting any assumptions on the control policy structure or objective.
As such, we find our methods to be highly complementary, and applicable at different stages of robot design. \par
One of the main challenges with viability-based approaches is tractability \cite{liniger2017real,bansal2017hamilton}. \rev{While we have shown how, in principle, a hierarchical control scheme reduces dimensionality, this approach alone is rarely sufficient in dealing with the curse of dimensionality on real systems.} There is much recent progress on different scalable approaches to computing viable and back-reachable sets (see Section \ref{sub:compvi}), and the specific choice will depend greatly on the properties of the system in question. \par
For running motion, characterized by nonlinear, non-smooth hybrid dynamics, we believe that, in addition to dimensionality reduction through hierarchical control, the use of heuristics such as computing ahead only two steps \cite{zaytsev2018boundaries}, are among the most promising tools to scaling this to real hardware. \par
We are also interested in using sampling-based approaches to make probabilistic estimates. There has been keen interest recently in applying machine learning techniques to tune control parameters directly in hardware \cite{antonova2016sample,heijmink2017learning,kumar2018improving,vonrohr2018gait}. In these situations, safe exploration of the state-action space is particularly important. Active sampling to add samples close to the edge of the viable set would significantly increase sample-efficiency for estimating the sets, while at the same time allowing safe exploration, making this a logical next step. \par
There is also potential for improvement in the definition of failures, the starting point of any viability approach. In this paper, we have used a very general and intuitive definition for failure (falling and direction reversal), however other definitions may be equivalent while offering earlier detection when computing viability kernels. Conservative definitions which lead to inner approximations may also be useful if they substantially speed up computation. It may also be possible to decouple the system dynamics, a common approach to simplifying control \cite{raibert1986legged, heim2016designing,whitman2009simple}, and identify different failure conditions for each decoupled subsystem. This divide and conquer approach would also allow substantially higher dimensional systems to be tackled.}
\section*{Appendix: SLIP and NSLIP models} \label{sec:appendix}

\begin{figure}[bt]
    \centering
    \includegraphics[width=0.95\columnwidth]{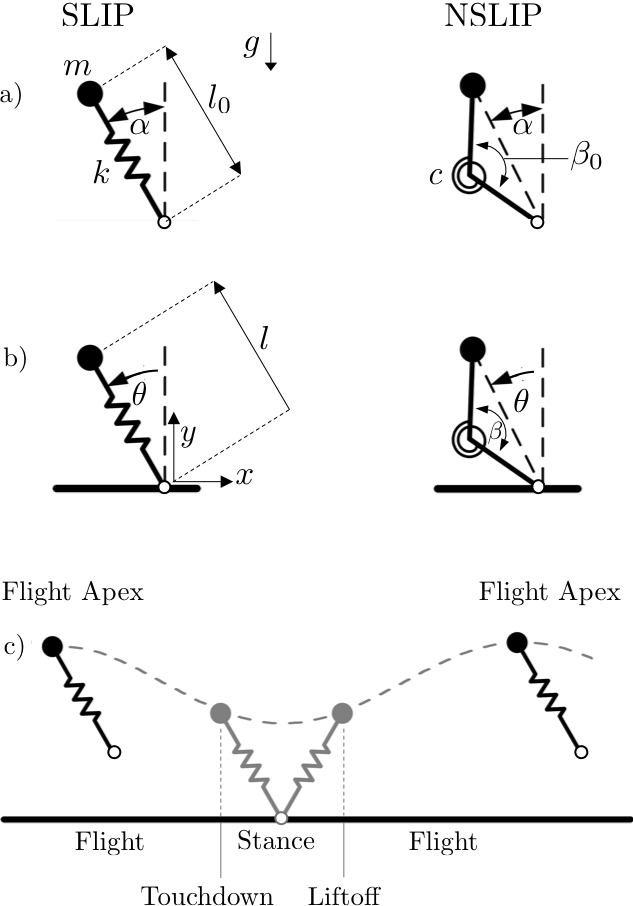}
    \caption{a) shows the parameters of the SLIP and NSLIP models. b) shows the states. The reference frame is reset to the foot position at each touchdown. c) shows a qualitative trajectory over a full cycle, with the relevant phases and events.} %
    \label{fig:fullslip}
\end{figure}

The SLIP and NSLIP models are shown in Fig. \ref{fig:fullslip}. Integration between two apex events is split into three phases: a flight phase which terminates with a touchdown event, a stance phase which terminates with a liftoff event, and another flight phase which terminates with an apex event.
The flight phase equations of motion are
\begin{equation*}
\left[
\begin{array}{c}
\ddot{x}\\
\ddot{y}\\
\end{array}
\right] = \left[
\begin{array}{c}
0\\
-g\\
\end{array}
\right]
\end{equation*}
where $x$ and $y$ are the body position and $g$ is the gravitational acceleration. The stance phase equations of motion are

\begin{align*}
\left[
\begin{array}{c}
\ddot{x}\\
\ddot{y}\\
\end{array}
\right] & = \frac{F_{leg}}{m}\left[
\begin{array}{c}
\sin\left(\theta\right)\\
\cos\left(\theta\right)\\
\end{array}
\right] - 
\left[
\begin{array}{c}
0 \\
g\\
\end{array}
\right] \\
\theta & = \arctan2\left(\frac{y}{x}\right) - \frac{\pi}{2}
\end{align*}
where $\theta$ is the incident angle between the body and the foot (the rotation by $\frac{\pi}{2}$ serves to keep it consistent with the landing angle of attack) and $F_{leg}$ is the force acting on the body due to the spring. In the SLIP model,

\begin{align*}
\text{SLIP: } F_{leg} & = k\left(l_0-l\right) \\
l & = \sqrt{(x^2 + y^2)}
\end{align*}
where $k$ is the spring coefficient, $l_0$ is the spring resting length, and $l$ is the leg length. In the NSLIP model,
\begin{align*}
\text{NSLIP: } F_{leg} & = \frac{4l c \left(\beta_0 - \beta\right)}{l_0^2 \sin \left(\beta \right)} \\
\beta & = \arccos \left(1-\frac{2l^2}{l_0^2}\right)
\end{align*}
where $c$ is the torsional spring coefficient, $\beta_0$ is the spring resting angle and $\beta$ is the knee angle.
The three events are
\begin{align*}
\text{touchdown: } & l = l_0 \\
\text{liftoff: } & \theta = \rev{\arctan2\left(\frac{y}{x}\right)} - \frac{\pi}{2} \\
\text{apex: } & \dot{y} = 0
\end{align*}
At each touchdown, the reference frame is reset to the foot position, which allows the equations of motion to be written more compactly. In the simulation, we also keep track of the foot position in an auxiliary variable. \par
For convenient comparison, we use the same parameters as in \cite{rummel2008stable}, which are similar to human averages: 
\begin{center}
\begin{tabular}{|l l l l|} 
\hline
gravitational acceleration &$g$ :& 9.81 &$\left[m/s^2\right]$ \\
\hline
body mass &$m$ :& 80 &$\left[kg\right]$ \\
\hline
prismatic spring resting length &$l_0$:& 1 &$[m]$ \\
\hline
prismatic spring coefficient &$k$ :& 8200 &$\left[N/m\right]$ \\ 
\hline
torsional spring resting angle &$\beta_0$ :& 170 &$\left[\degree\right]$\\
 \hline
torsional spring coefficient &$c$ :& 704 &$\left[Nm/rad\right]$ \\
\hline
\end{tabular}
\end{center}
For the SLIP and NSLIP simulations shown, except in Fig. \ref{fig:stepdown}, the system energy simulated is 1'860 Joules.
\pagebreak[1]
\section*{ACKNOWLEDGMENTS}
We thank everyone who gave feedback during the writing of this manuscript. In particular, we appreciate the frequent and insightful discussions with Matthew Millard, Brent Gillespie and Andrea del Prete, as well as Friedrich Solowjow's advice on mathematical notation. We also appreciate the editors and reviewers for their constructive suggestions and quick turnaround time. \\

\printbibliography

\begin{IEEEbiography}[{\includegraphics[width=1in,height=1.25in,clip,keepaspectratio]{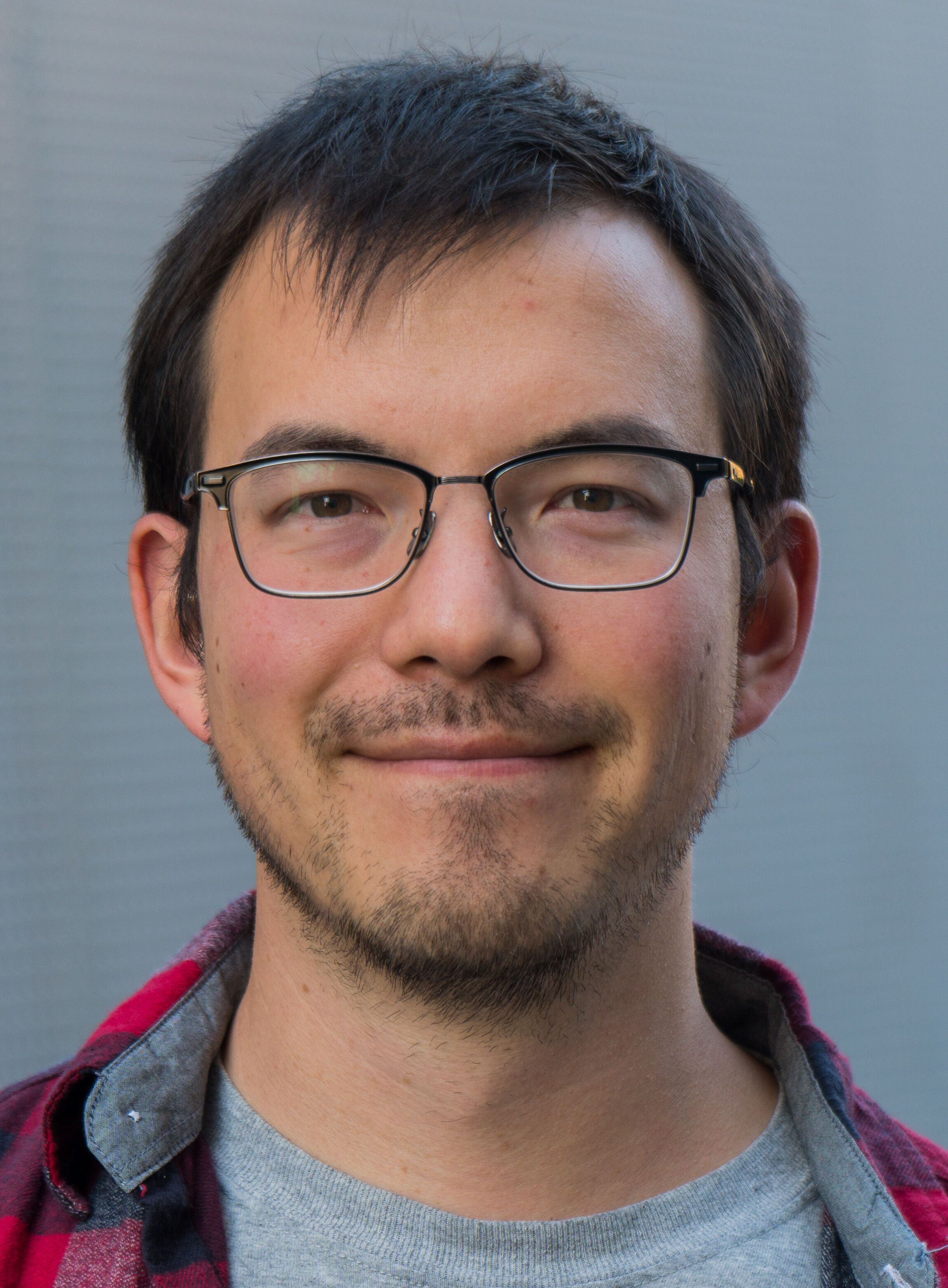}}]{Steve Heim}
received the B.Sc. degree in mechanical engineering and M.Sc. in robotics, systems and control from ETH Zurich, Switzerland, in 2012 and 2015, respectively. He then spent two years with the Ishiguro lab at Tohoku University in Sendai, Japan. He is currently pursuing the Ph.D. degree with the Dynamic Locomotion Group at the Max Planck Institute for Intelligent Systems in Stuttgart, Germany. His research interests include nonlinear dynamics, control and learning, particularly in relation to legged locomotion.
\end{IEEEbiography}
\begin{IEEEbiography}[{\includegraphics[width=1in,height=1.25in,clip,keepaspectratio]{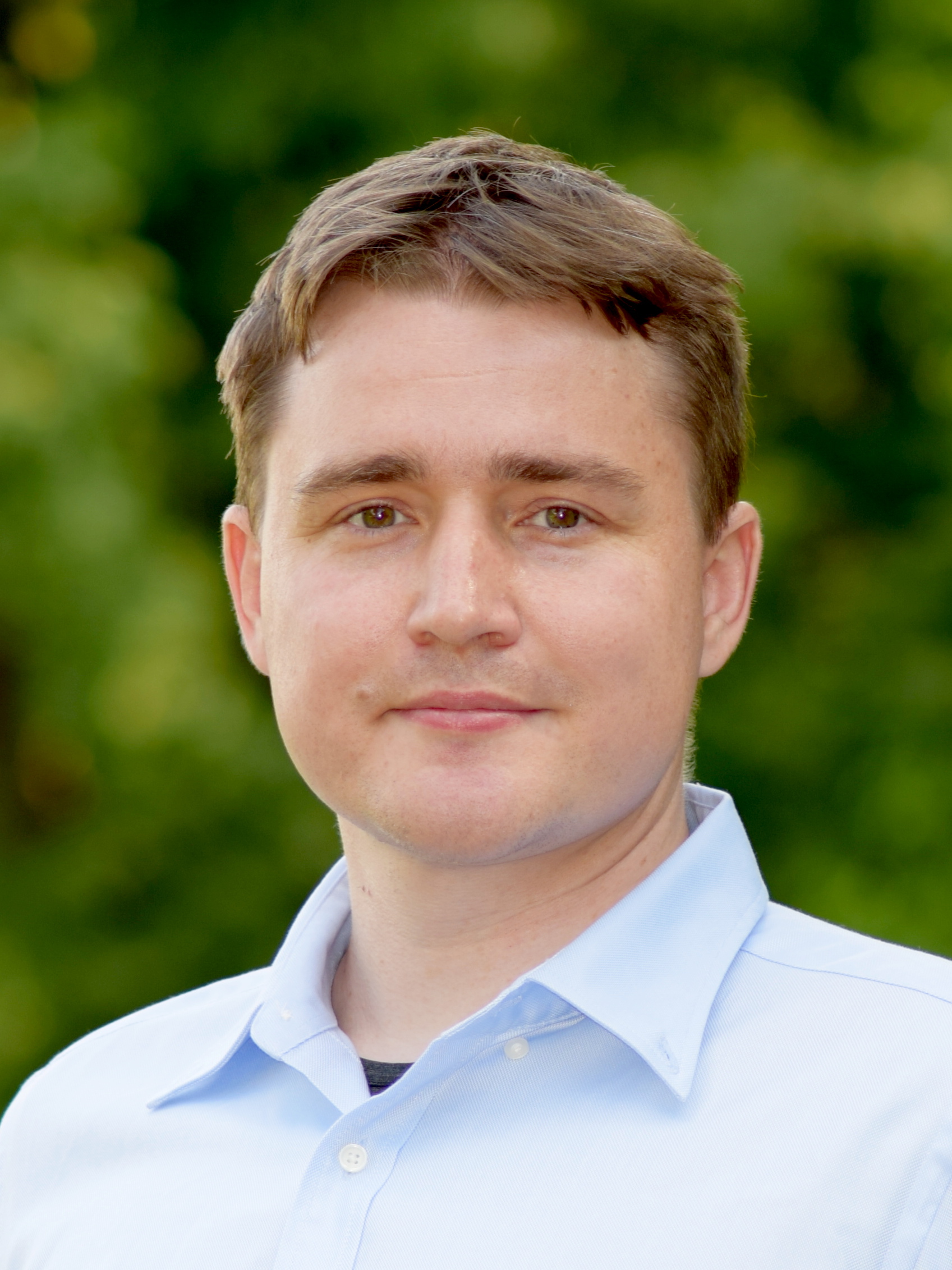}}]{Alexander Spr{\"o}witz}
received the Diplom degree in mechatronics from Ilmenau Technical University in Germany in 2005, and the Ph.D. degree in manufacturing systems and robotics from the Biorobotics Laboratory at the Swiss Federal Institute of Technology in Lausanne (EPFL), Switzerland in 2010.
Since 2016 he is the Max Planck Research Group Leader of the Dynamic Locomotion Group, and IMPRS-IS faculty at the Max Planck Institute for Intelligent Systems in Stuttgart, Germany. 
His current research focuses on the mechanisms underlying legged locomotion. Dr. Spröwitz and his team design and experiment with legged robots and models to infer biomechanics and neuro-control principles of motion in animals.
\end{IEEEbiography}

\end{document}